\documentclass[journal]{IEEEtran}
\usepackage{graphicx}
\newcommand{\fref}[1]{Fig. \ref{#1}}
\newcommand{\sref}[1]{Section \ref{#1}}
\newcommand{\tref}[1]{TABLE \ref{#1}}

\usepackage{array}
\usepackage{amsmath}
\usepackage{bm}

\usepackage{algorithm}
\usepackage{algorithmic}

\usepackage{array}
\usepackage{multirow}
\usepackage{booktabs}
\usepackage{graphicx}
\ifCLASSOPTIONcompsoc
\usepackage[caption=false, font=normalsize, labelfont=sf, textfont=sf]{subfig}
\else
\usepackage[caption=false, font=footnotesize]{subfig}
\fi
\usepackage{url}
\usepackage{color,xcolor}
\begin{document}

\title{Deep Reinforcement Learning for Combinatorial Optimization: Covering Salesman Problems}
%
%
%

\author{Kaiwen~Li,
        Tao~Zhang,
        Rui~Wang
        Yuheng~Wang,
        and Yi~Han
\thanks{This paper is partially supported by the National Natural Science Foundation of China (No. 72071205 and No. 61773390).}  

\thanks{Kaiwen Li, Tao~Zhang and Rui~Wang (corresponding author) are with the College of Systems Engineering, National University of Defense Technology, Changsha 410073, PR China, and also with the Hunan Key Laboratory of Multi-Energy System Intelligent Interconnection Technology, HKL-MSI2T,
Changsha 410073, PR China. (e-mail: kaiwenli\_nudt@foxmail.com, zhangtao@nudt.edu.cn, ruiwangnudt@gmail.com)}

\thanks{Yuheng Wang is with the Graduate College, National University of Defense Technology, Changsha 410073, PR China}
\thanks{Yi Han is with the Science and Technology on Parallel and Distributed Processing Laboratory, College of Computer, National University of Defense Technology, Changsha 410073, PR China}
\thanks{Manuscript received April 19, 2021; revised August 26, 2021.}}

%
%

\markboth{Journal of IEEE Transactions on Cybernetics, ~Vol.~14, No.~8, February~2021}%
{Shell \MakeLowercase{\textit{et al.}}: Bare Demo of IEEEtran.cls for IEEE Journals}

\maketitle

\begin{abstract}
This paper introduces a new deep learning approach to approximately solve the Covering Salesman Problem (CSP). In this approach, given the city locations of a CSP as input, a deep neural network model is designed to directly output the solution. It is trained using the deep reinforcement learning without supervision. Specifically, in the model, we apply the Multi-head Attention to capture the structural patterns, and design a dynamic embedding to handle the dynamic patterns of the problem. Once the model is trained, it can generalize to various types of CSP tasks (different sizes and topologies) with no need of re-training. Through controlled experiments, the proposed approach shows desirable time complexity: it runs more than 20 times faster than the traditional heuristic solvers with a tiny gap of optimality. Moreover, it significantly outperforms the current state-of-the-art deep learning approaches for combinatorial optimization in the aspect of both training and inference. In comparison with traditional solvers, this approach is highly desirable for most of the challenging tasks in practice that are usually large-scale and require quick decisions.
\end{abstract}


\IEEEpeerreviewmaketitle

\section{Introduction}

The Traveling Salesman Problem (TSP) is a frequently studied combinatorial optimization problems in the field of operation research. Given a set of cities with their spatial locations, the goal of TSP is to find a minimum length tour that visits each city once and returns back to the original city. In this work we focus on the Covering Salesman Problem (CSP), which is a generalization of the TSP. In the CSP, each city is given a predefined covering distance, within which all other cities are covered. The CSP seeks a minimum length tour over a subset of the given cities such that each city has to be either visited or has to be covered by at least one city on the tour. The CSP reduces to a TSP if the covering distance of each city is zero. Thus the CSP is NP-hard and is more difficult to be solved than the TSP. In some practical scenarios, due to the limitation of fuel or manpower resources, it is hard to guarantee that each city can be travelled exactly once as assumed in the TSP. Hereby, the CSP arises in various and heterogeneous real-world applications such as emergency management and disaster planning \cite{current1989covering,shariff2012location,reina2013evolutionary}. For example, in the routing problem of healthcare delivery teams \cite{current1989covering}, it is not necessary to visit each village, because the villages that are not visited are expected to go to their nearest stop for service. 

Traditional approaches to solving such NP-hard combinatorial optimization problems mainly include three categories \cite{vesselinova2020learning}: exact algorithms, approximate algorithms and heuristic algorithms. Exact algorithms can produce optimal solutions by enumeration or other techniques such as branch-and-bound, however, require forbidding computing time when tackling large instances. Approximate algorithms, in general, can obtain near-optimal solutions in polynomial time with theoretical optimality guarantees. However, such approximate algorithms might not exist for all types of optimization problems and the execution time can be still prohibitive if their time complexity is higher-order polynomial. Heuristic methods are more favorable in practice since they usually run faster than the above two types of approaches. But they lack theoretical guarantees for the quality of the solutions. In addition, as iteration-based approaches, heuristic methods still suffer from the limitation of long computation time when tackling with large instances as a large number of iterations are required for population updating or heuristic searching. Moreover, esoteric domain knowledge and trial-and-error are usually required when designing such heuristic methods, leading to considerable designing effort and time. 

As most of the challenging problems in real-world applications are large-scale and are usually under the constraint of execution time, traditional algorithms suffer from specific limitations when applied to practical challenging tasks: forbidding computation time and the need to be revised or re-executed whenever a change of the problem occurs. This can be impractical for large-scale tasks in real-world applications. Recent advances in deep learning have shown promising ability of solving NP-hard decision problems. A well-known example is the inspiring success of employing deep reinforcement learning to solve the game Go \cite{silver2017mastering}, which is a complex discrete decision problem. In the context of the advances attained by the deep learning in solving various decision tasks, recently, some works have focused on using the deep learning to solve classical NP-hard combinatorial problems, including the TSP. They replace the carefully handcrafted heuristics by the policy parameterized by a deep neural network and learn the policy from data. They provide a new paradigm for combinatorial optimization: the solution is directly output by a deep neural network which is trained on a collection of instances from a certain type of problem. Such learning-based approaches have shown appealing advantages over traditional solvers in the aspect of computational complexity and the ability of scaling to unseen problems \cite{vinyals2015pointer, kool2018attention}. 

In literature, most of studies focuses on designing different types of heuristic algorithms to tackle CSP. These carefully designed heuristics can certainly improve the performance, however, are problem-specific and suffer from the aforementioned limitations. In this paper, we propose a new deep learning approach to approximately solve CSP. The promising idea to leverage the deep learning for combinatorial optimization has been tested on TSP. However, as a generalization of the TSP, the CSP appears harder to be addressed due to its dynamic feature. Therefore we propose a powerful deep neural network model based on Multi-head Attention and dynamic embedding that can effectively map from the problem input of the CSP to its solution. The model is trained using deep reinforcement learning in an unsupervised way. The proposed approach has empirically shown a handful of merits in the following aspects:

\begin{itemize}
\item \textbf{Optimality.} On solving the CSP, the proposed approach significantly outperforms the recent state-of-the-art deep learning approaches for combinatorial optimization. 
\item \textbf{Execution time.} The presented approach runs more than 20 times faster than the traditional heuristic solvers with only a small optimality gap. It offers a desirable trade-off between the execution time and the optimality of solutions.
\item \textbf{Scalability.} Once trained, the model can be used to solve, not only one, but a collection of CSP instances of different size and/or different city locations as long as they are from the same data generating function. It can scale to unseen instances with no need of re-training.

\item \textbf{Generalization.} The proposed approach can also generalize to different CSP tasks, e.g., CSPs where each city has a fixed covering radius within which all cities are covered, or CSPs where each city can cover a fixed number of cities. The model that is only trained on one type of CSP task, however, can solve different types of CSP tasks without the need to re-train the model. 
\end{itemize}

\section{Related work}
\subsection{Covering Salesman Problems}
The covering salesman problem (CSP) is first formulated in 1989 by Current and Schilling \cite{current1989covering}. They presented a straightforward heuristic approach to solve it. It can be divided into two parts. First a minimum number of cities that can cover all the cities are selected, namely a set covering problem (SCP). The second phase can be regarded as a TSP: the shortest tour is found on the above selected cities. Since usually more than one solution can be found for the SCP, the TSP solver is applied on all the found solutions and the TSP solution with the minimum length is output as the CSP solution.

From then on, a number of good heuristics are designed to solve the CSP effectively. Golden et al. \cite{golden2012generalized} proposed two local search algorithms (LS1 and LS2) to solve CSP. LS1 and LS2 have been widely used as benchmark approaches in the literature. As local search methods, they all start from a random feasible solution and improve it by making use of different operators such as destroy, repair and permutation. LS1 first removes a fixed number nodes from the current solution according to a deletion probability. New nodes are then inserted into the current tour one by one according to a insertion probability until the solution is feasible again. These probabilities are determined by the decrease or increase of the tour length when deleting or adding that node into the tour. In contrast, LS2 removes one node a time from the current solution and then insert the nearest nodes from the removed node into the solution. Lin-Kernighan Procedure and 2-Opt procedure is also used to improve the solution. 

Salari et al. \cite{salari2012integer} developed an Integer Linear Programming (ILP) based heuristic method to solve the CSP. Its main difference from the method of \cite{golden2012generalized} is its idea of applying ILP to improve the solution. This approach first applies the similar destroy and repair operations to decrease the tour length by removing some nodes from the current solution and inserting new nodes back to make a feasible solution. The solution is further improved by solving an ILP model with the objective of minimizing the tour length. Other local search approaches also apply the similar destroy and repair operations to solve the CSP. For example, Shaelaie et al. \cite{shaelaie2014generalized} proposed a variable neighborhood search method and Venkatesh et al. \cite{venkatesh2019multi} proposed a multi-start iterated local search algorithm.

In addition to the local search approaches, various population based heuristic approaches have been proposed. Salari et al. \cite{salari2015combining} combines ant colony optimization (ACO) algorithm and dynamic programming to tackle the CSP. It also incorporates various local search procedures such as removal, insertion and the 3-OPT search. Experiments show this approach to be superior on large size instances. Tripathy et al. \cite{tripathy2017metameric} presented a genetic algorithm (GA) with new designed crossover operators for solving the CSP. However its performance fails to defeat LS1 and LS2. Different types of GAs \cite{pandiri2020two, shaelaie2014generalized} have also been developed to solve the CSP. Additionally, Pandiri et al. \cite{pandiri2019artificial} proposed an artificial bee colony algorithm for this problem. 

In terms of the performance, the above heuristic algorithms achieve a same level of optimality to the earlier proposed LS1/LS2 on most instances. Some of the heuristics may outperform LS1/LS2 on several instances. For years, various heuristics have been designed to solve CSP with significant specialized knowledge and trial-and-error efforts. However, there was no significant breakthrough and currently no machine learning methods have been proposed for the CSP to the best of our knowledge.

\subsection{Deep Learning Approaches for Combinatorial Optimization}

The first application of using Neural Networks (NNs) to tackle the challenge of combinatorial optimization is the Hopfield-Network \cite{hopfield1985neural} proposed in 1985. However, it is trained only to solve one instance a time with little advantages over traditional solvers. 

Recent five years have seen a surge in the applications of deep learning for combinatorial optimization. Deep learning approaches for combinatorial optimization are usually based on the \emph{end-to-end} learning mode, that is, using a Deep Neural Network (DNN) to directly output the optimal solution. Current state-of-the-art approaches mainly use \emph{sequence-to-sequence} networks \cite{sutskever2014sequence} and Graph Neural Networks (GNNs) \cite{scarselli2008graph} for combinatorial optimization. Representative works are reviewed as follows.

Vinyals et al. \cite{vinyals2015pointer} first proposed to use a sequence-to-sequence model for combinatorial optimization, also known as the  \emph{Pointer Network}. It uses the attention mechanism to output a permutation of an input sequence. The model is trained offline using pairs of TSP instances and their (near) optimal tours in a supervised fashion. It successfully solves the small size TSP and reinvigorates this line of work that applies deep learning for combinatorial optimization. 

Learning from supervised labels might be inapplicable because it is expensive to construct high-quality labeled data and it prohibits the learned model from performing better than the traning examples. In this context, Bello et al. \cite{bello2016neural} proposed to use an Actor-Critic \cite{konda2000actor} deep reinforcement learning (DRL) algorithm to train the Pointer Network in an unsupervised manner. It takes each TSP instance as a training sample and uses the tour length of the solution obtained by the current policy as the reward, which is used to update the policy parameters via the policy gradient formula. This approach achieves a comparable performance to \cite{vinyals2015pointer} on small instances and can further solve larger TSP instances (up to 100 cities). Nazari et al. \cite{nazari2018deep} replaces the LSTM \cite{hochreiter1997long} encoder of the Pointer Network by a simple node embedding. It can save up to 60\% training time for the TSP while maintaining a similar performance. This model can also solve the Vehicle Routing Problem (VRP) by adding additional dynamic elements to the attention mechanism. 

Khalil et al. \cite{khalil2017learning} considered the graph structure of the combinatorial optimization problem by adopting a \emph{structure2vec} GNN model. It consecutively inserts a node into the current partial solution according to the node scores parameterized by \emph{structure2vec}. The model is trained using the DQN method \cite{mnih2013playing} and is applied on the Minimum Vertex Cover and Maximum Cut problems other than the TSP. Mittal et al. \cite{mittal2019learning} replaced the \emph{structure2vec} of \cite{khalil2017learning} by a Graph Convolutional Network (GCN) and followed the same greedy algorithm. It performs better than \cite{khalil2017learning} on large graphs.

Nowak et al. \cite{nowak2017note} also used a GNN to model the problem. But it outputs an adjacency matrix instead, which is then converted into a feasible solution by beam searching. The model is trained with supervision. As this method constructs solutions in a non-autoregressive manner, it performs worse than the above autoregressive approaches for the TSP. Joshi et al. \cite{joshi2019efficient} followed the same paradigm to \cite{nowak2017note} but used a more effective GCN to encode the problem instance. Experiments show its superior performance on the TSP when using beam search and shortest tour heuristic to construct the solutions, however, it requires much longer execution time. Moreover, Li et al. \cite{li2018combinatorial} used a GCN to output the probability map that represents the likelihood of each node belong to the optimal solution. Consequently a guided tree search is used to construct the solution according to the probability map of all nodes. 

Inspired by the Transformer architecture \cite{vaswani2017attention}, which is the state-of-the-art model in the field of seqence-to-seqence learning, authors in \cite{deudon2018learning, kool2018attention} applied the Multi-head Attention mechanism of the Transformer to tackle the combinatorial optimization challenge. They build the TSP solutions autoregressively using the attention similar to the Pointer Network. Deudon et al. \cite{deudon2018learning} demonstrated that incorporating a 2-OPT local search \cite{croes1958method} can improve the performance. Kool et al. \cite{kool2018attention} designed a more effective decoder and proposed to train the model using a new reinforcement learning method with a greedy rollout baseline. This approach achieves the state-of-the-art performance among the concurrent deep learning approaches for the TSP. Moreover, it can scale to a variety of practical problems like the VRP, the Orienteering Problem, etc.

In addition, Li et al. \cite{li2020deep} adopted the deep learning approaches for the multi-objective TSP. This approach outperforms traditional multi-objective optimization solvers in terms of execution time and solution quality. Other combinatorial optimization problems tackled by similar deep learning models include the Maximal Independent Set \cite{abe2019solving}, the Graph Coloring \cite{yolcu2019learning}, the Boolean Satisfiability \cite{yolcu2019learning}, etc. Recently, Joshi et al. \cite{joshi2019learning} explored the impact of different training paradigms and revealed favorable properties of reinforcement learning over supervised learning.

\section{Preliminaries}
\subsection{Problem Definition}
In the CSP, we are given a complete graph $G=(V,E)$ where $N=\{1,2,\ldots,n\}$ is the node set that represents the cities. $E=(\{i, j\}: i, j \in N, i<j)$ is the edge set and $c_{ij}$ represents the cost of edge $\{i,j\}$ which is usually defined as the shortest distance between node $i$ and $j$. Each node $i$ can cover a subset of nodes $S_i$. In the CSP, we are expected to find a Hamiltonian tour over a subset of nodes $V$ such that cities that are not on the tour must be covered by at least one city on the tour. The objective is to minimize the total tour length. 

In this paper, we focus on the Euclidean CSP: the cost of edge $\{i,j\}$ is defined by the Euclidean distance between node $i$ and $j$ given their two-dimensional geographical locations.

\subsection{Attention Theory for Combinatorial Optimization}
\label{basic}
For ease of understanding our model in this paper, we first introduce the fundamental theory of how to use deep learning techniques to solve the combinatorial optimization problem, the TSP specifically. 

\textbf{Node embedding.} \quad The initial \emph{node embedding} of node $i$ is a mapping of the 2-dimensional input of node $i$, i.e., two-dimensional coordinates of its location, to a $d_h$-dimensional vector ($d_h=128$ in this work).

\begin{figure}[htbp]
	\centering
	\includegraphics[width=1\linewidth]{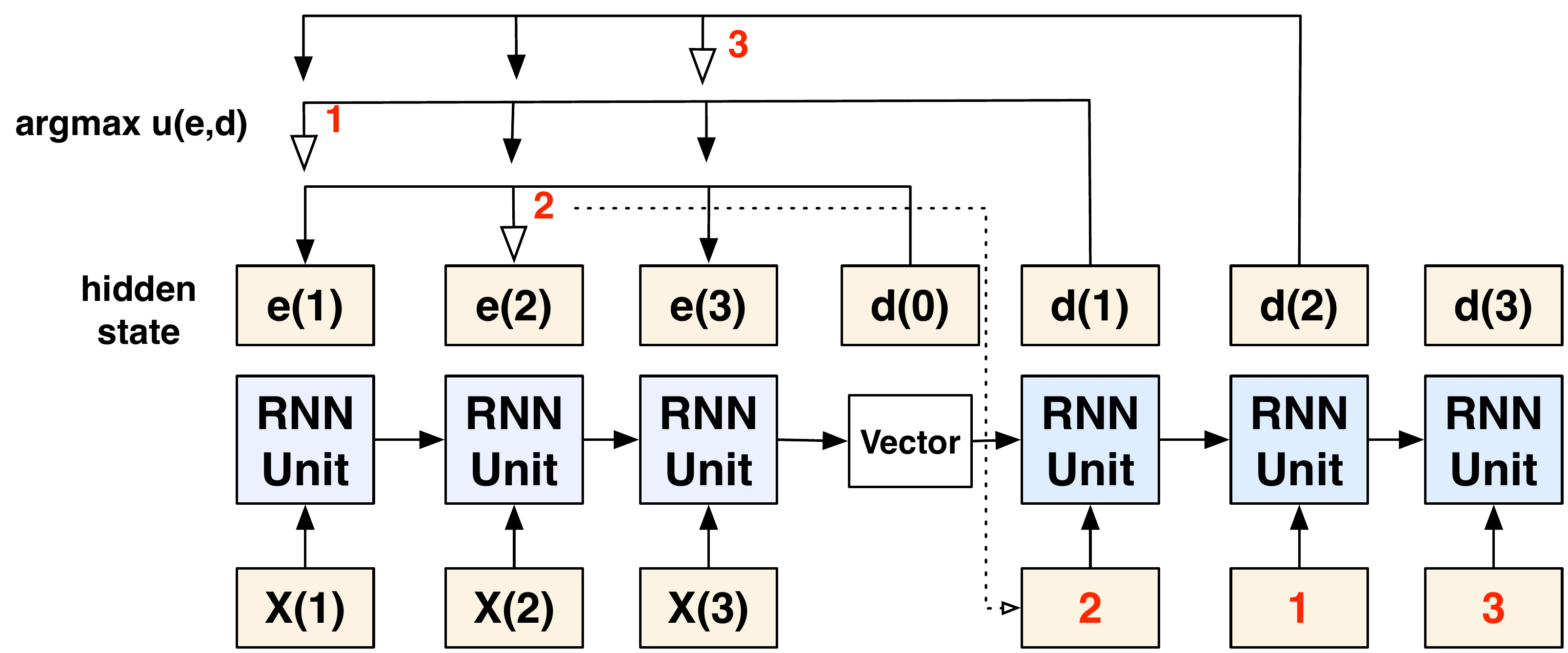}
	\caption{Take a 3-city TSP instance as an example to demonstrate the Pointer Network. Inputs are the embeddings of the city locations. The cities are selected step by step. At each step $t$, the city with the largest $u^t$ is selected. $u_i^t$ for each city $i$ is calculated according to \eqref{eq:pointer}.}
	\label{fig:pointer}
\end{figure}

\textbf{Pointer Network.} \quad The model architecture of Pointer Network is shown in \fref{fig:pointer}. Its basic structure is the Encoder-Decoder. Encoder is used to encode the node embeddings to a \emph{Vector}, while decoder is to decode the \emph{Vector} into an output sequence. Either the encoder or decoder is a Recurrent Neural Network (RNN). While using the encoder RNN to encode the inputs, the hidden state $e_i$ is produced for each city $i$. Here, $e_i$ can be interpreted as the feature vector of city $i$. The final hidden state of the encoder RNN is the \emph{Vector}. 

As indicated in \fref{fig:pointer}, the \emph{Vector} is used as input to the decoder RNN, and the decoding happens sequentially. The \emph{Vector} also serves as the initial hidden state $d_0$, which is used to select the first city to visit. The next hidden state $d_1$ is obtained by inputting the node embedding of the selected city to the encoder RNN. Specifically, at decoding step $t$, the probability of city selection is calculated by the encoder state $d_{t-1}$ and the feature vector $e_i$ of each city. The RNN reads the node embedding of the selection and output the current encoder state $d_{t}$, which is then used in the next decoding step. The calculation of the probability is namely the Attention method and is computed as follows. 

\begin{equation}
\begin{array}
{c}{u_{i}^{t}=v^{T} \tanh \left(W_{1} e_{i}+W_{2} d_{t}\right) \quad i \in(1, \ldots, n)} \\ 
{P\left(y_{t+1} | y_{1}, \ldots, y_{t}, X \right)=\operatorname{softmax}\left(u^{t}\right)}
\end{array}
\label{eq:pointer}
\end{equation}
where $v, W_1, W_2$ are learnable parameters. The one with the largest probability can be selected as the city to be visited at step $t$. 

For example, as shown in \fref{fig:pointer}, to select the first city at decoding step $t=1$, $u_{1}^{1}, u_{2}^{1}, u_{3}^{1}$ are calculated by $(d_0, e_1)$, $(d_0, e_2)$ and $(d_0, e_3)$ according to Equation \eqref{eq:pointer}. The one with the largest $u^1$, i.e., $u_{2}^{1}$, is selected as the first city, which is city 2. By inputting the node embedding of city 2 into the decoder RNN, we can get $d_1$. The next city is selected according to the probabilities calculated by applying the attention formula on $(d_1, e_1)$, $(d_1, e_2)$ and $(d_1, e_3)$. Then the procedure loops till all the cities are selected. After training the parameters $v, W_1, W_2$ offline, the Pointer Network obtains the ability to output the desired tour of the cities.

\textbf{Self-attention.} \quad In the above Pointer Network, RNN is used to get the feature vector $e_i$ of each city $i$ in the encoder. Self-attention achieves the same functionality. It is the basic component of Transformer\cite{vaswani2017attention}, which is the state-of-the-art model in the field of sequence-to-sequence learning like machine translation. By applying self-attention in the encoder, the obtained attention value of each input $i$ stores not only the information of itself, but also its relations with other inputs. Thus self-Attention can be more effective in capturing the structural patterns of the problem. 

Recall the attention mechanism used in \eqref{eq:pointer}, $d_t$ in the $t_{th}$ city selecting step can be seen as a \emph{query}. It contains the information of already visited cities. The representation $e_i$ of each city $i$ can be seen as a \emph{key}. The $i_{th}$ city is given more \emph{Attention} if its \emph{key} is more compatible with the \emph{query}. In \eqref{eq:pointer}, the \emph{Attention} value is interpreted as the probability of being selected. 

Self-attention works by calculating the attention value of each component in a sequence to all other components. And the obtained attention value of the $i_{th}$ component serves as its representation. Therefore the processed embeddings contain more information compared to its original sequence. Different from the attention calculation method in \eqref{eq:pointer}, the \emph{Scaled Dot-Product Attention} is used in Transformer's self-attention: the compatibility of query $\mathbf{q}_t$ and key $\mathbf{k}_i$ is simply calculated as their multiply $\mathbf{q}_t^T\mathbf{k}_i$ and then scaled by a scaling factor. The self-attention mechanism is introduced as follows:

The basic attention is realized by \emph{query} and the \emph{key-value} pair, that is, more attention would be given to \emph{value} if its \emph{key} is more compatible with \emph{query}. For self-attention in the encoder, \emph{query}, \emph{key} and \emph{value} are all linear projections of the input node embeddings:

\begin{equation}
\mathbf{Q}=W^{Q} \mathbf{X}, \quad \mathbf{K}=W^{K} \mathbf{X}, \quad \mathbf{V}=W^{V} \mathbf{X}.
\label{qkv}
\end{equation}

That is, the \emph{query}, \emph{key} and \emph{value} of $i_{th}$ node is:
\begin{equation}
\mathbf{q}_{i}=W^{Q} \mathbf{X}_{i}, \quad \mathbf{k}_{i}=W^{K} \mathbf{X}_{i}, \quad \mathbf{v}_{i}=W^{V} \mathbf{X}_{i}.
\end{equation}

\begin{figure}[htbp]
	\centering
	\includegraphics[width=0.7\linewidth]{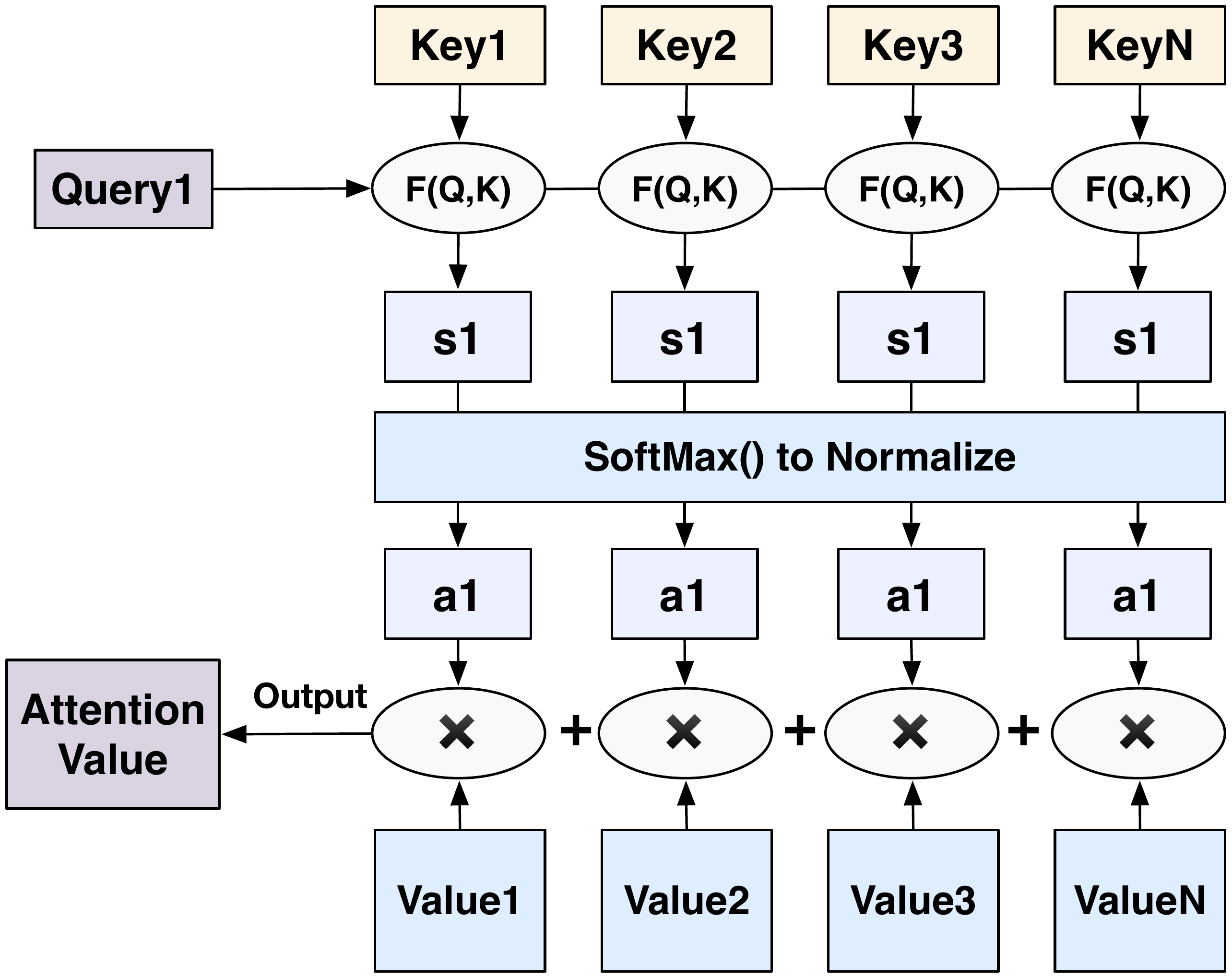}
	\caption{The process of calculating the attention value given the \emph{query} and \emph{key-value} pair.}
	\label{fig:self-attention}
\end{figure}

The process of calculating the attention value of the $i_{th}$ node is depicted in \fref{fig:self-attention}. First the relations of the $i_{th}$ node with other nodes are computed by a compatibility function of the \emph{query} of the $i_{th}$ node with the \emph{keys} of all other nodes: $\mathbf{q}_i\mathbf{K}^T$. The obtained compatibility serves as weights and are normalized by the softmax function. The attention value is finally computed as a weighted sum of the \emph{values} $\mathbf{V}$, where the weights are $\mathbf{q}_i\mathbf{K}^T$. Thus, the attention value of the $i_{th}$ node is $\mathbf{q}_i\mathbf{K}^T\mathbf{V}$, which has the same dimension ($d_h$) with its initial node embedding $\mathbf{X}_{i}$. As a result of the self-attention process, the computed attention value of the $i_{th}$ node stores not only its own knowledge but also its relations with other nodes. 

In practice, we compute the attention values of all nodes simultaneously by matrix multiplying:
\begin{equation}
\text { Attention }(Q, K, V)=\operatorname{softmax}\left(\frac{Q K^{T}}{\sqrt{d_{k}}}\right) V
\end{equation}
where $\sqrt{d_{k}}$ is the scaling factor and $d_k$ is the dimension of the \emph{key}.

\textbf{Multi-Head Attention (MHA).} \quad As pointed out in \cite{vaswani2017attention}, it is beneficial to carry out the attention process multiple times and obtain $M$ attention values. Then the $M$ attention values are linearly projected into a final attention value. Hence, various features can be extracted using the MHA process. 

Overall, structural patterns of the problem can be extracted effectively by using the MHA in the encoder. 

\section{Attention Model}
\label{model}

We still follow the \emph{Encoder-Decoder} architecture to model the CSP. The encoder is used to construct the \emph{key} of the model that consists of two parts: static embeddings obtained by the MHA that contain the structural patterns of the problem; dynamic embeddings obtained by a designed guidance vector that describes the dynamic patterns of the problem. The decoder is designed with a RNN and an additional attention operator.

The model is formally described as follows. Taking a CSP instance $s$ with $n$ cities as an example, the feature $\mathbf{x}_i$ of city $i$ is defined by its location, i.e., its x-coordinate and y-coordinate. A solution $\bm{\pi}=\left(\pi_{1}, \dots, \pi_{k}, k \leq n \right)$ is a permutation of a subset of the cities. All the cities should be visited or covered by traveling along the tour $\bm{\pi}$. With this formulation of CSP, the Attention-dynamic model is designed. The model takes city locations of instance $s$ as input, and outputs the sequence $\bm{\pi}$. More formally, the model defines a policy $p(\bm{\pi}|s)$ for selecting $\bm{\pi}$ given $s$: 

\begin{equation}
p_{\theta}(\boldsymbol{\pi} | s)=\prod_{t=1}^{k} p_{\boldsymbol{\theta}}\left(\pi_{t} | s, \boldsymbol{\pi}_{1 \sim t-1}\right), k \leq n.
\end{equation}

The encoder encodes the original features $\mathbf{x}$ and produces the final embeddings (representations) of the cities. The decoder takes as input the encoder embeddings, summarizing the information of previously selected cities $\boldsymbol{\pi}_{1 \sim t-1}$ and then outputs $\pi_t$, one city at a time. $\boldsymbol{\theta}$ represents the model parameters such as $W^{Q}$, $W^{K}$ and $W^{V}$. With an  optimal set of $\boldsymbol{\theta}^*$, the model has the ability of outputting the optimal tour $\bm{\pi}^*$ given a problem instance $s$.

\subsection{Encoder}
The purpose of the encoder is to produce a representation for each city, that is, produce the \emph{key} of each city. Our \emph{key} consists of two parts: static embeddings and dynamic embeddings.

\begin{figure*}[!t]
	\centering
	\includegraphics[width=0.7\linewidth]{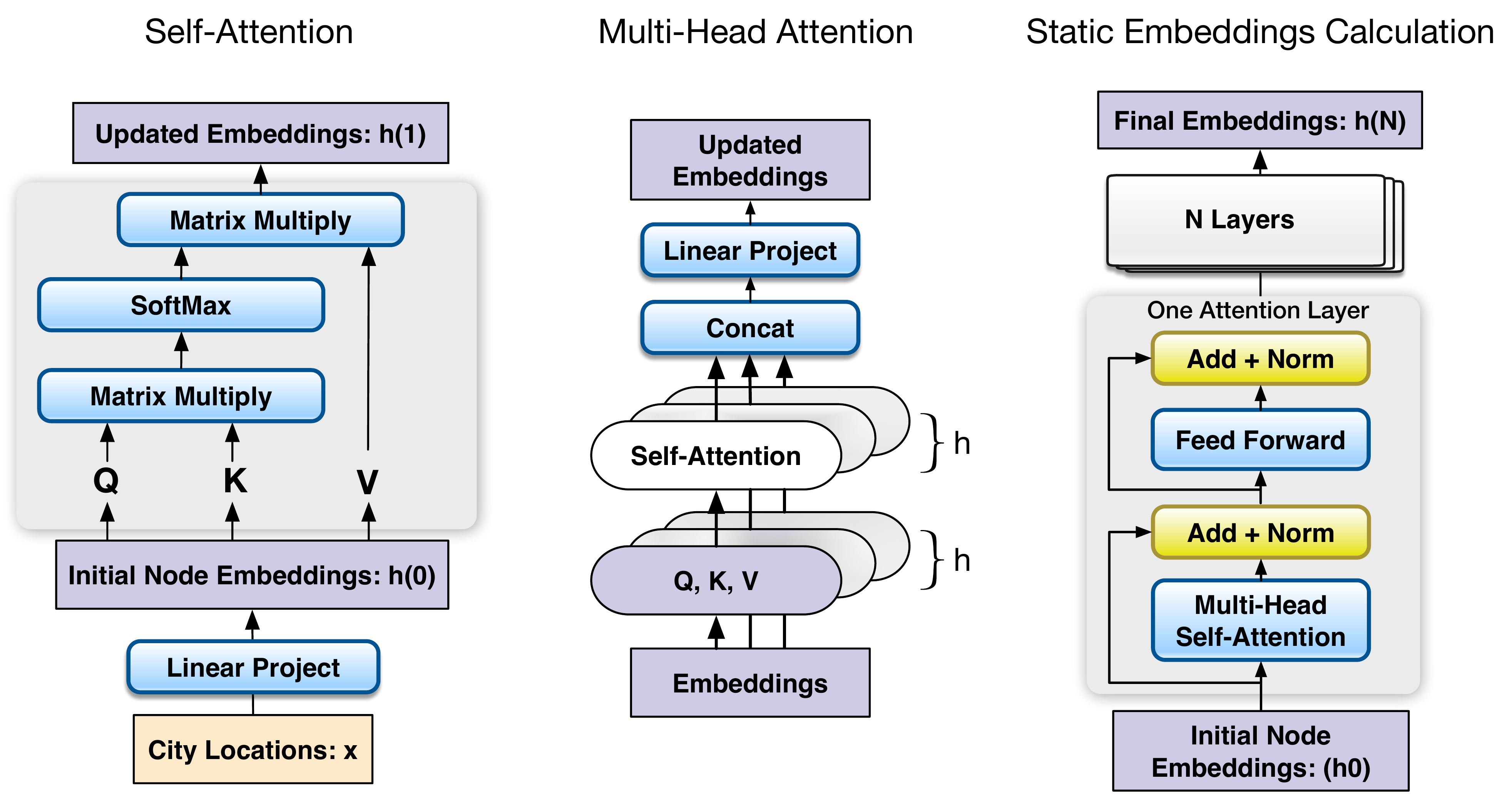}
	\caption{This figure gives the process of calculating the static embeddings in our model. It consists of $N$ layers of multi-head attention modules which is formed by the self-attention mechanism.}
	\label{fig:model}
\end{figure*}

\textbf{Construction of static embeddings.} \quad The static embedding demonstrate the static feature of a city, i.e., its location. It is constructed similar to the TSP task in \cite{deudon2018learning, kool2018attention}. We employ the introduced MHA to produce the static embeddings, as shown in \fref{fig:model}. Its basic element is the \emph{Self-attention}, which can produce the representation for each city $i$ that stores not only its own feature but also its relations with other cities. As depicted in \fref{fig:model}, in the MHA, the self-attention is conducted $h$ times and the obtained $h$ embeddings are projected into a final embedding.

The architecture of computing the static embeddings consists of $N$ sequentially connected \emph{attention layers} that all have the same structure. That is, the embeddings are processed $N$ times sequentially. Each attention layer has two sub-layers: the first is a MHA layer, and the second is a fully connected feed-forward layer. In addition, each sub-layer is added with a residual connection \cite{he2016deep} and a layer normalization. Thus, the output of each sub-layer becomes $Norm(x+Sublayer(x))$. 

Overall, computing static embeddings needs two steps: 
\begin{itemize}
\item We first map the $d_x$-dimensional city locations ($d_x$ = 2) to the $d_h$-dimensional node embeddings ($d_h$ = 128): 

\begin{equation}
\mathbf{h}_{i}^{(0)}=W^{\mathbf{x}} \mathbf{x}_{i}+\mathbf{b}^{\mathbf{x}}.
\end{equation}

$W^{\mathbf{x}}$ and $\mathbf{b}^{\mathbf{x}}$ are learnable parameters. Assuming there are $n$ cities, thus the shape of the initial node embeddings is $n \times d_h$. 
\item As shown in \fref{fig:model}, the initial node embeddings $h^{(0)}$ are processed and updated through $N$ Attention layers. The output embeddings of the $\ell_{th}$ layer are computed as:
\begin{equation}
\begin{aligned} \\\mathbf{h}_{tmp} &=Norm^{\ell}\left(\mathbf{h}^{(\ell-1)}+\mathrm{MHA}^{\ell}\left(\mathbf{h}^{(\ell-1)}\right)\right) \\ 
\mathbf{h}^{(\ell)} &=Norm^{\ell}
\left(\mathbf{h}_{tmp}+\mathrm{FF}^{\ell}\left(\mathbf{h}_{tmp}\right)\right) \end{aligned}.
\end{equation}

The output embedding $h^{(N)}$ of the $N_{th}$ layer is the final static embedding, where $h^{(N)}_i$ is the $d_h$-dimensional static feature vector of city $i$.
\end{itemize}

\textbf{Construction of dynamic embeddings.} \quad 
The static embeddings $h^{(N)}$ can be directly taken as the \emph{key} for the TSP \cite{kool2018attention}. However, the states of the cities in a CSP are not static during the decoding step. For example, once a city is covered, its possibility of being visited might be reduced. And the possibility can be rather reduced if it is covered more times. Thus it is vital to introduce the concept of \emph{Covering} into the construction of \emph{key} for CSP optimization. Aiming at this, we define a guidance vector $g_i$ for each city $i$ that indicates its state of covering, and $g_i$ is dynamically updated at each decoding step of $t$.

\quad Assuming city $\pi_t$ is selected to be visited at decoding step $t$, ${C(\pi_t)}$ is defined as the set of cities that are covered by $\pi_t$. And the number of nodes in ${C(\pi_t)}$ is counted as $N_{C(\pi_t)}$. Cities in ${C(\pi_t)}$ are then sorted according to their distances to $\pi_t$. Thus, each city in ${C(\pi_t)}$ has a sorted index $c_i$ ranging from 1 to $N_{C(\pi_t)}$, where $c_i=1$ indicates that city $i$ is nearest to $\pi_t$, and vice versa. 

\quad At first, $g_i$ is initialized to 1 for all cities: 
\begin{equation}
g_i=1, i=1,\cdots,n.
\end{equation}
And at each decoding step $t$, $g_i$ is updated dynamically:
\begin{equation}
g_i=g_i \times \frac{c_i}{N_{C(\pi_t)}}, i \in C(\pi_t), i=1,\cdots,n.
\label{eq:guidance}
\end{equation}
According to \eqref{eq:guidance}, the dynamic state of city $i$ is represented by a value $g_i$. Supposing that city $\pi_t$ is visited, the nearest city to $\pi_t$ is assigned with the smallest value. $g_i$ is further reduced if city $i$ is covered again by other cities. 

Hence, the obtained guidance vector contains the covering information for each city: the state that it is covered or not and its distance to the one who covers it. It can thus guide the selection of the next city, e.g., the one that is closer to the already visited cities may have a lower probability of being selected. 

\quad  It is noted that we do not explicitly reduce the selection probability of city $i$ according its $g_i$ value. It is because that a covered city can also be a good option in some occasions. Instead, we let the model itself to learn how to utilize $g_i$ to adjust the probability. Aiming at this, $g_i$ is linearly projected to a $d_h$-dimensional embedding $G_i$ by learnable weights $W_G$:

\begin{equation}
G_i = g_i W_G
\end{equation}

As $G_i$ dynamically changes during the decoding, $G_i$ is called the dynamic embedding in this study. While the static embedding $h^{(N)}$ indicates the static states of the cities, i.e., their spatial locations, the dynamic embedding $G$ indicates their dynamic states of \emph{covering}. Then we construct the $key$ according to $h^{(N)}$ and $G$:

\begin{equation}
\text k_i = W^Kh_i^{(N)}G_i, i=1 \cdots n
\end{equation}
where $\mathbf{k}_i$ represents the $key$ for city $i$ and $W^K$ are the learnable weights with size $(d_h \times d_h) $. In this way, the model can capture both of the static and dynamic patterns of the problem and thus learns better policies.

\subsection{Decoder}

The decoder is composed of two parts: RNN, whose output $d_t$ is used to compute the \emph{query} $q_t$ at decoding step $t$ ; Encoder-Decoder Attention, which is used to calculate the probability of city selections according to \emph{query} $q_t$ and \emph{key}. Intuitively, the city with the largest probability is selected as $\pi_t$ at decoding step $t$. The probability is calculated through the Encoder-Decoder Attention based on the \emph{query} $q_t$ from the RNN and the \emph{key} from the encoder. That is, among all $key_i, i=1 \ldots n$, the one that most matches $q_t$ is selected as the next city to be visited. This part illustrates how to build the \emph{query} in the model.

\textbf{Construction of \emph{query}.} \quad To select the next city to be visited, it is naturally to take the previously visited cities as the \emph{query}. As RNN is capable of handling such sequential information, in this study, GRU \cite{cho2014learning}, which is a variant of RNN, is employed to construct the \emph{query} in the decoder. We set the mean of the encoder embeddings $\overline{\mathbf{h}}=\frac{1}{n} \sum_{i=1}^{n} \mathbf{h}_{i}^{(N)}$ as the initial hidden state of the decoder RNN, i.e., $d_0 = \overline{\mathbf{h}}$. Meanwhile, we use the learned parameter $\mathbf{v}_1$ as the first input to the decoder RNN at step $t=1$. In the following steps $t>1$, the input to the RNN is the node embedding $\mathbf{h}_{\pi_{t-1}}^{(N)}$ of the previous selection $\pi_{t-1}$:

\begin{equation}
o_t, d_t = f_{GRU}\left(  I_t, d_{t-1} \right), d_0 = \overline{\mathbf{h}}
\end{equation}
\begin{equation}
I_t=\left\{\begin{array}{ll}{
\mathbf{v}_1} & {t=1} \\ 
{\mathbf{h}_{\pi_{t-1}}^{(N)}} & {t>1}

\end{array}\right.
\end{equation}

$d_t$ is the hidden state of the decoder RNN at decoding step $t$, and it stores the information of previously selected cities $\pi_{1 \sim t-1}$. As the selection of $\pi_{t}$ largely depends on $\pi_{1 \sim t-1}$, $d_t$ can be employed as the \emph{query} naturally. 

\quad However, $\pi_{1 \sim t-1}$ is not the only factor that determines $\pi_{t}$. To construct the \emph{query}, the relations between the already visited cites and the rest of cities should also be considered. For example, the city that is closer to the last visited city might have more opportunity to be visited next. Thus we further process $d_t$ using the Attention mechanism as follows:

\begin{equation}
\text q_t=Attention(d_t, K_1, V_1)=\operatorname{softmax}\left(\frac{d_t K_1^{T}}{\sqrt{d_{k}}}\right) V_1,
\end{equation}
where $K_1$ and $V_1$ are linear projections of the node embeddings $h^{(N)}$ from the encoder. Thus the \emph{query} $q_t$ for selecting $\pi_t$ is defined as the attention value of $d_t$ and the key-value pairs $K_1, V_1$. By adding the extra attention operation to $d_t$, the obtained \emph{query} contains richer information: the compatibility of previously selected cities (represented by $d_t$) with all other cities (represented by $h^{(N)}$).

\subsection{Compute Probability by Attention}

At the decoding step $t$, we compute the selection probability of all the cities $i=1 \cdots n$ based on the \emph{query-key} $(q_t,k_i)$ by the Scaled Dot-Product Attention:

\begin{equation}
u_i^t =\frac{\mathbf{q}_t^T\mathbf{k}_i}{\sqrt{d_k}},
\end{equation}

where $\sqrt{d_k}$ is the scaling factor and $d_k = \frac{d_h}{M}$. $M$ is the number of heads in the MHA. In addition, we mask the cities that are already visited to make sure they will not be selected again:

\begin{equation}
u_i^t=\left\{\begin{array}{ll}
-\infty & \text { if } j \in  \pi_{1 \sim t-1}
\\
\frac{\mathbf{q}_t^{T} \mathbf{k}_i}{\sqrt{d_{k}}} & \text { otherwise. }
\end{array}\right.
\label{mask}
\end{equation}

The probability will be zero after applying softmax operator to the already visited cities with $u_i^t = -\infty$. The final probability is computed by scaling $u_i^t$ using the softmax operator:

\begin{equation}
p_{i}=\frac{e^{u_i^t}}{\sum_{j=1}^n e^{u_j^t}}, i=1, \cdots, n
\label{softmax}
\end{equation}

Assuming that the greedy strategy is used, the city with the largest probability $p_i$ is selected to be visited at decoding step $t$. 

The model is therefore composed of the above encoder, decoder and attention. The encoder takes as input the features of the CSP instance and outputs the encoder embeddings. Decoding happens sequentially from $t=0$ and stops until all the cities are visited or covered. The selected cities forms the final solution $(\pi_0, \pi_1, \cdots )$, i.e., permutation of the cities.

\section{Train with REINFORCE Algorithm}

Given a CSP instance $s$, Section \ref{model} defines the model parameterized by parameters $\boldsymbol{\theta}$ that can produce the probability distribution $p_{\boldsymbol{\theta}}(\boldsymbol{\pi} | s)$ by Equation \eqref{mask} and \eqref{softmax}. The solution $\boldsymbol{\pi}|s$ can be obtained by sampling from $p_{\boldsymbol{\theta}}(\boldsymbol{\pi} | s)$. Since the goal is to minimize the total length $L(\pi)$ of the cyclic tour, it is suitable to train the model with the REINFORCE algorithm \cite{Williams1998Simple}, which uses Monte Carlo rollout to compute the rewards, i.e. play through the whole episode to find a complete cyclic tour and the total tour length is computed as the reward (-$L(\pi)$). Then the agent can learn to improve itself by the \emph{state-action-reward} tuple.

Formally, given the actions $\boldsymbol{\pi}|s$ and the corresponding reward/loss $L(\pi)$, parameters $\boldsymbol{\theta}$ of the model can be updated by gradient descent using the REINFORCE algorithm \cite{Williams1998Simple}: 
\begin{equation}
\begin{aligned}
\nabla_{\boldsymbol{\theta}} \mathcal{L}(\boldsymbol{\theta}) & = \mathbf{E}
_{p_{\boldsymbol{\theta}}(\boldsymbol{\pi} | s)}
\left[ \nabla \log p_{\boldsymbol{\theta}}(\boldsymbol{\pi} | s)L(\boldsymbol{\pi})\right]
\\
\boldsymbol{\theta} &  \leftarrow \boldsymbol{\theta} +\nabla_{\boldsymbol{\theta}} \mathcal{L}(\boldsymbol{\theta}).
\end{aligned}
\label{PG}
\end{equation}

Since a number of city selection actions are sampled from the probability distribution and the total length is computed as the rewards, the recorded rewards of different episodes have high variance due to the uncertainty of the sampling. The variance provides conflicting descent direction for the model to learn and hurt the speed of convergence. To reduce the variance, it is common to introduce the baseline $b(s)$ to rewrite the policy gradient in \eqref{PG}:
\begin{equation}
\nabla_{\boldsymbol{\theta}} \mathcal{L}(\boldsymbol{\theta}) = \mathbf{E}
_{p_{\boldsymbol{\theta}}(\boldsymbol{\pi} | s)}
\left[ \nabla \log p_{\boldsymbol{\theta}}(\boldsymbol{\pi} | s)(L(\boldsymbol{\pi})-b(s))\right].
\end{equation}

Loosely speaking, $b(s)$ serves as the average performance. A policy is encouraged if its action performs better than the average. A popular method is to train an additional neural network, i.e., the critic network, to approximate the $b(s)$. Its parameters are learned via pairs of $(s, L(\boldsymbol{\pi}))$. 

However, as reported in \cite{kool2018attention}, training a policy network and a critic network simultaneously is hard to converge. Therefore we follow the greedy rollout baseline as introduced in \cite{kool2018attention} which is proved to be superior than the critic baseline in the traditional Actor-Critic training algorithm. 

Here, no additional neural network is required to approximate $b(s)$. Instead, given a CSP instance $s$, $b(s)$ is computed as the length of the tour from a deterministic greedy rollout of the baseline policy $p_{\boldsymbol{\theta}^{*}}$. The greedy rollout of a policy means that a solution is built by running the policy greedily, i.e., at each step, the city with the largest output probability is selected. The baseline policy is the best model so far during training. At the end of each training epoch, once the trained model is better than the baseline policy, we replace the baseline policy with the current model. 

Algorithm \ref{alg:train} outlines the REINFORCE training algorithm with the greedy rollout baseline and Adam optimizer \cite{kingma2014adam}. In this way, if a sampled solution $\boldsymbol{\pi}$ is better than the greedy rollout of the best model ever, $(L(\boldsymbol{\pi})-b(s))$ is negative and thus reinforcing such actions, and vice versa. Thus, the model can effectively learn to solve the problem, i.e., minimizing the total tour length.

\begin{algorithm}[htb]
\caption{Training Algorithm using REINFORCE}
\label{alg:train}
\begin{algorithmic}[1]
\REQUIRE {training set $\mathcal{X}$, batch size $B$, number of epochs, $N_{epoch}$, number of steps per epoch $N_{steps}$}
\STATE Initialize model parameters of the current policy and baseline policy $\boldsymbol{\theta}, \boldsymbol{\theta}^*\leftarrow\boldsymbol{\theta}$

\FOR{$epoch \leftarrow 1:N_{epoch}$}
\FOR{$step \leftarrow 1:N_{step}$}
\STATE $s_{i} \leftarrow \operatorname{RandomInstance}(\mathcal{X}) \text { for } i \in\{1, \ldots, B\}$
\STATE $\boldsymbol{\pi}_{i} \leftarrow  p_{\boldsymbol{\theta}}(s_{i}).$ Run the policy by sampling
\STATE $\boldsymbol{\pi}_{i}^* \leftarrow  p_{\boldsymbol{\theta}^*}(s_{i}).$ Run the policy greedily
\STATE $\nabla \mathcal{L} \leftarrow \sum_{i=1}^{B}\left(L\left(\boldsymbol{\pi}_{i}\right)-L\left(\boldsymbol{\pi}_{i}^{*}\right)\right) \nabla_{\boldsymbol{\theta}} \log p_{\boldsymbol{\theta}}\left(\boldsymbol{\pi}_{i}\right) $
\STATE Update $\boldsymbol{\theta}$ by $\operatorname{Adam}$ according to $\nabla \mathcal{L}$
\ENDFOR
\IF{$p_{\boldsymbol{\theta}}$ performs better than $p_{\boldsymbol{\theta}^*}$}
\STATE $\boldsymbol{\theta}^*\leftarrow\boldsymbol{\theta}$
\ENDIF
\ENDFOR

\end{algorithmic}
\end{algorithm}

\section{Experiment Setup}

We investigate the performance of the proposed approach on 20-, 50-, 100-, 200-, and 300-CSP tasks. 

\textbf{Baseline Algorithms.}\quad The presented approach (Attention-dynamic) is compared to both of the recent state-of-the-art deep learning techniques as well as classical non-learned heuristic methods. The most popular deep learning methods on solving combinatorial optimization problems in the past few years are included in the comparisons:

\begin{itemize}
\item Pointer Network \cite{vinyals2015pointer}. (PN) 
\item Pointer Network with dynamic embeddings \cite{nazari2018deep}. (PN-dynamic) 
\item Attention model \cite{kool2018attention}. (Attention)
\end{itemize}

The PN model \cite{vinyals2015pointer} is the first deep learning model that can effectively tackle combinatorial optimization problems. It is a widely used baseline approach. The PN-dynamic model \cite{nazari2018deep} aims at solving the VRP problem whose state changes during the decision process that has the similar dynamic feature with CSP. Attention model \cite{kool2018attention} is the current state-of-the-art deep learning method for solving various combinatorial optimization problems. Thus these two models are included as competitors. 

Before the emergence of deep learning techniques, heuristic methods are the main approaches to solve CSP. Therefore, in addition to deep learning methods, we also include non-learned heuristic methods for comparisons. LS1 and LS2 \cite{golden2012generalized} are commonly used benchmark algorithms for solving CSP from the literature \cite{golden2012generalized, shaelaie2014generalized, salari2012integer, salari2015combining}. Since all the deep learning models run on Python platform and no source code is found for LS1/LS2, we implemented LS1/LS2 in Python to make fair comparisons. The code is written strictly according to \cite{golden2012generalized} and is publicly available\footnote{https://github.com/kevin031060/CSP\_Attention/tree/master/LS}.

\textbf{Hyperparameters Configurations.} \quad 
Hyperparameters used for training our model are shown in \tref{table:param}. Across all experiments of our model and the compared neural combinatorial baselines, we use mini-batches of 256 instances, 128-dimensional node embeddings and Encoder-Decoder with 128 hidden units. Models are trained using the Adam optimizer \cite{kingma2014adam} with a fixed learning rate $10^{-4}$. 

In the training phase, 320,000 CSP instances are generated and used in training for 50 epochs. The training CSP instances are randomly generated: city locations are randomly chosen from a uniform distribution that ranges from [0,1]$\times$[0,1]; the number of nearest neighbor cities that can be covered by each city (NC) is set as 7. It is worthy noting that NC is set to a fixed number upon training the model, however, it can be set as an arbitrary number when testing. This validates the generalization ability of the model and will be  further discussed in \sref{general}. We use the same set of hyperparameters and the same training dataset with fixed random seed for training other neural combinatorial baseline models. As for LS1 and LS2, we follow the recommended hyperparameters from \cite{golden2012generalized} for different problem dimensions.

\begin{table}
\caption{Hyperparameters Configurations}
\begin{tabular}{cc|cc}
\hline 
HyperParameters & Value & HyperParameters & Value \\ 
\hline 
Batch size & 256 & Hidden dimension & 128 \\ 
\hline 
No. of epoch & 50 & No. of heads & 8 \\ 
\hline 
No. of instances per epoch & 320000 & No. of Encoder layers & 3 \\ 
\hline 
Optimizer & Adam & Learning rate & 1e-4 \\ 
\hline 
\end{tabular} 
\label{table:param}
\end{table}

\textbf{Implementations.} \quad
All experiments are implemented in Python and conducted on the same machine with one GTX 2080Ti GPU and Intel 64GB 16-Core i7-9800X CPU. Our code is open source \footnote{https://github.com/kevin031060/CSP\_Attention} to reproduce the experimental results and to facilitate future studies.

\textbf{Evaluation Procedure.} \quad
We evaluate the performance on held-out test set of randomly generated 100 CSP instances and take the average predicted tour length as the performance indicator. Since the compared LS1 and LS2 require long execution time, we only generate 100 instances for testing. As pointed out in \cite{deudon2018learning}, a posterior local search can effectively improve the quality of solutions obtained by the neural combinatorial model. Thus we conduct a simple local search to further process the obtained tour of our model. The running time of our method is the sum of the inference time of the neural network model and the execution time of the local search. 

\section{Results and Discussion}
The learning, approximation and generalization ability of the presented approach are demonstrated through the following experimental results. 

\subsection{Comparisons against deep learning baselines on training}

We first evaluate the learning ability of our method in the training phase against the recent state-of-the-art deep learning baselines: PN, PN-dynamic and Attention model.

\begin{figure}[tbph]
\centering
\includegraphics[width=3in]{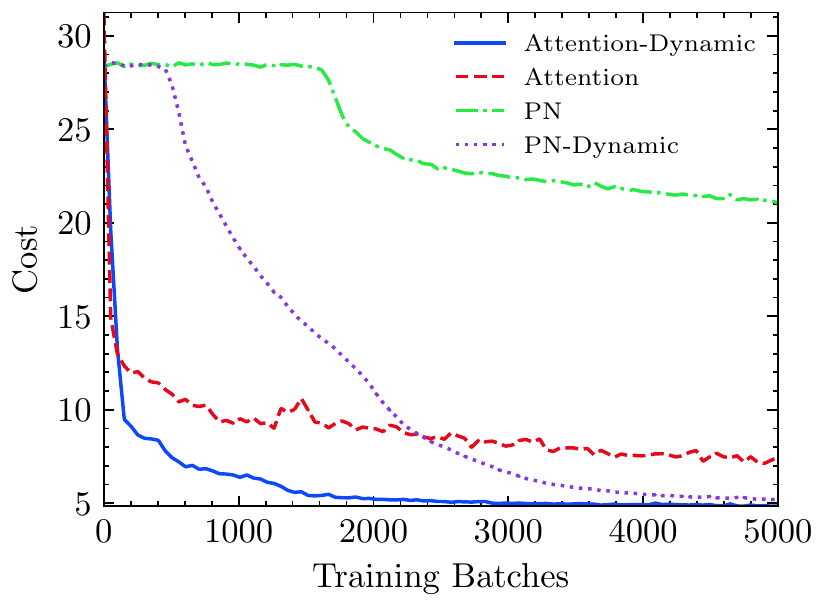}
\caption{Held-out validation set cost as a function of the number of training batches for the compared models.}
\label{fig:training}
\end{figure}

\begin{figure}[tbph]
\centering
\includegraphics[width=3in]{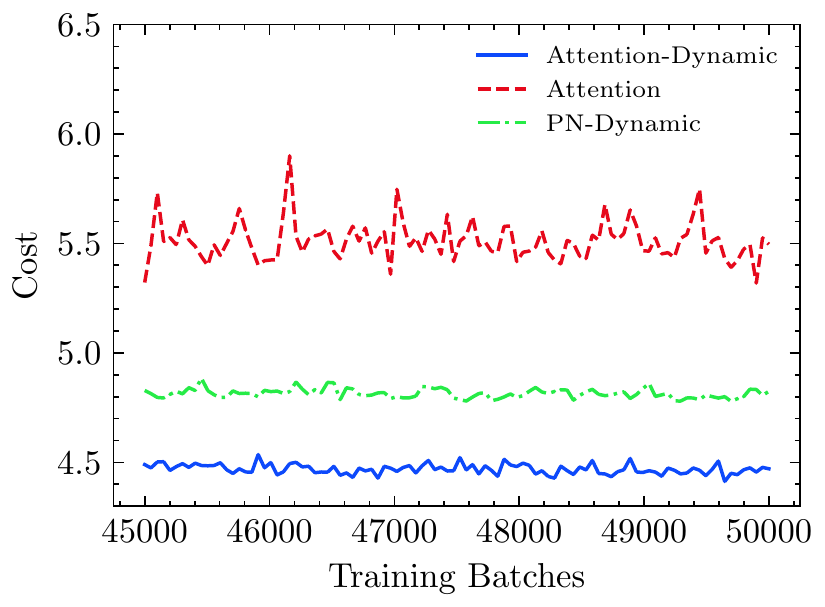}
\caption{The final costs of PN-dynamic, Attention, Attention-dynamic at the end of the training phase.}
\label{fig:training_end}
\end{figure}

\fref{fig:training} compares the performances of the compared approaches during training. The average predicted tour length of the held-out validation set is taken as the cost to evaluate the models in the training phase. As the lines may be too close, \fref{fig:training_end} shows the final cost of the compared models at the end of the training phase.

 We observe that our method clearly outperforms the compared baselines in terms of both the convergence speed and the final cost. PN and Attention model fail to converge to the optimum as they do not take into account the dynamic information of the covering sate of the cities. Although PN-dynamic model considers the dynamic features, it is defeated by our model since the MHA mechanism that we used can effectively extract the feature of the CSP task thus helps converge significantly faster and lead to a lower cost than the PN-dynamic model. Our approach appears to be more stable and sample efficient in comparison with the baselines. 

\begin{table}[t!]
  \centering
  \caption{Comparisons of training time for 100 batches(in seconds).}
    \begin{tabular}{l|cc}
       Method   & \multicolumn{1}{l}{CSP50} & \multicolumn{1}{l}{CSP100} \\
    \midrule
    \midrule
    PN    & 85.6  & 176.2 \\
    PN-dynamic & 107.5 & 237.8 \\
    AM    & 26.3  & 45.9 \\
    AM-dynamic & 32.6  & 52.6 \\
    \bottomrule
    \bottomrule
    \end{tabular}%
  \label{tab:train_time}%
\end{table}%

Additionally, \tref{tab:train_time} outlines the training time of the compared models. Despite the low performance of the Pointer Network-based models, they appear to be computational inefficient during training. Our method can save up to 75\% of the training time while achieving a lower cost than the PN-dynamic model.

\subsection{Model performance on test set}

To evaluate the model performances, both the small-scale CSP tasks (CSP20, CSP50, CSP100) and the larger-scale tasks (CSP150, CSP200, CSP300) are used for testing. We mainly focus on learning from small-scale training instances and measure the generalization to larger and unseen test instances: the model trained on CSP20 instances is used to solve randomly generated CSP20 and CSP50 test problems; the model trained on CSP100 instances is used to solve CSP100 and CSP150 test problems; and the model trained on CSP200 instances is used to solve CSP200 and CSP300 test problems. We empirically find that the performances deteriorate significantly for all deep learning methods when using model trained on CSP20 to solve CSP300 instances. Therefore, to fairly evaluate the model performance, we do not use such experimental setup with a large gap of the instance size between training and testing. We use the average predicted tour length as the cost to evaluate the model. And the optimality gap of the approaches w.r.t. the best solver is also used as the performance metric. 

\tref{tab:small} and \tref{tab:large} presents the performances of the compared approaches on small scale CSP tasks (CSP20, CSP50, CSP100) and larger scale CSP tasks (CSP150, CSP200, CSP300) respectively. Results of the non-learned heuristic methods are shown in the first part of the tables while the second part shows the results obtained by the pure deep learning models without a posterior local search. 

\textbf{Performances over deep learning baselines.} It is obvious that our method clearly outperforms all deep learning baselines on all CSP tasks. Current state-of-the-art deep learning methods on solving combinatorial problems fails to solve the CSP task, leading to a large optimality gap w.r.t. the classical heuristic solvers. Our method comfortably surpasses these deep learning baselines with a much smaller gap to the heuristic solvers. 

\textbf{Performances over heuristic baselines.} We then conduct a simple local search to improve the solutions obtained by our end-to-end model. Results are shown in the third part of \tref{tab:small} and \tref{tab:large}. It can be seen that additionally conducting a local search can improve the solutions, however, leading to a longer execution time. This observation is consistent with the results reported in \cite{deudon2018learning}. The hybrid method in basis of our attention-dynamic model can achieve significantly closer optimality (no more than 2\% on all tasks except CSP20) to the traditional heuristic method like LS1. Although there is still an optimality gap, our method runs dramatically faster than the heuristic methods. It is about 20 to 40 times faster than the LS1\&LS2 with the local search, and thousands times faster than LS1\&LS2 without local search. Note that our approach is trained on a different dataset from the test instances. The results are obtained by generalizing the model to unseen CSP tasks with different city locations and dimensions. Thus it is reasonable that there is a slight performance gap between our method and the heuristic solvers. Also it is not our goal to defeat the specialized, carefully designed heuristic solvers. Instead, we aim to show the fast solving speed and high generalization ability of the proposed approach.

It should be noticed that the heuristic solvers usually require a stop searching criterion, e.g., the cost not changing for a fixed number of epochs. Therefore, with extra searching procedures, their execution time can be longer than expected. The above experiments are conducted with the heuristic methods optimized to their best. However, it would be more interested to see the comparison of running time with all the methods executed to the same level of optimality. 

In this context, we additionally design controlled experiments to make the comparisons fair and explicit, i.e., comparing their execution time by stopping them at the same cost. In specific, the cost obtained by our attention-dynamic(LS) method is set as the benchmark cost and the heuristic methods immediately stop once they reach or surpass the defined cost. Their execution time when stopping at the same cost is recorded. Results are shown in \tref{tab:stop}. Clearly, our approach is more than 10 times faster than the traditional heuristic solvers if they all reach the same level of optimality. It shows the significantly favorable time complexity of our approach than the traditional solvers.

\begin{table*}[htbp]
  \centering
  \caption{Attention-dynamic method vs baselines on small-scale CSPs. Results of the cost and the execution time are recorded in average by evaluating the models on 100 randomly generated instances. The gap is computed w.r.t. the best performing approach.}
    \begin{tabular}{l|rrr|rrr|rrr}
    \multicolumn{1}{c|}{\multirow{2}[3]{*}{Method}} & \multicolumn{3}{c|}{CSP20} & \multicolumn{3}{c|}{CSP50} & \multicolumn{3}{c}{CSP100} \\
\cmidrule{2-10}          & \multicolumn{1}{l}{Cost} & \multicolumn{1}{l}{Gap} & \multicolumn{1}{l|}{Time/s} & \multicolumn{1}{l}{Cost} & \multicolumn{1}{l}{Gap} & \multicolumn{1}{l|}{Time/s} & \multicolumn{1}{l}{Cost} & \multicolumn{1}{l}{Gap} & \multicolumn{1}{l}{Time/s} \\
    \midrule
    \midrule
    LS1   & 1.87  & 8.09\% & 4.24  & 2.57  & 0.00\% & 15.04 & 3.58  & 0.00\% & 88.85 \\
    LS2   & 1.73  & 0.00\% & 6.55  & 2.67  & 3.89\% & 21.61 & 3.69  & 3.07\% & 107.98 \\
    \midrule
    PN    & 4.27  & 146.82\% & 0.019 & 7.85  & 205.45\% & 0.027 & 12.05 & 236.59\% & 0.042 \\
    PN-dynamic & 2.15  & 24.28\% & \textbf{0.011} & 3.29  & 28.02\% & \textbf{0.015} & 4.68  & 30.73\% & \textbf{0.033} \\
    AM    & 2.23  & 28.90\% & 0.014 & 3.87  & 50.58\% & 0.041 & 5.02  & 40.22\% & 0.064 \\
    AM-dynamic & \textbf{1.89} & \textbf{9.25\%} & 0.014 & \textbf{2.75} & \textbf{7.00\%} & 0.032 & \textbf{4.08} & \textbf{13.97\%} & 0.063 \\
    \midrule
    AM-dynamic (LS) & \textbf{1.85} & \textbf{6.94\%} & \textbf{0.5} & \textbf{2.64} & \textbf{2.72\%} & \textbf{0.71} & \textbf{3.65} & \textbf{1.96\%} & \textbf{2.66} \\
    \bottomrule
    \bottomrule
    \end{tabular}%
  \label{tab:small}%
\end{table*}%

\begin{table*}[htbp]
  \centering
  \caption{Attention-dynamic method vs baselines on large-scale CSPs. Results of the cost and the execution time are recorded in average by evaluating the models on 100 randomly generated instances. The gap is computed w.r.t. the best performing approach.}
    \begin{tabular}{l|rrr|rrr|rrr}
    \multicolumn{1}{c|}{\multirow{2}[3]{*}{Method}} & \multicolumn{3}{c|}{150-TSP} & \multicolumn{3}{c|}{200-TSP} & \multicolumn{3}{c}{300-TSP} \\
\cmidrule{2-10}          & \multicolumn{1}{l}{Cost} & \multicolumn{1}{l}{Gap} & \multicolumn{1}{l|}{Time/s} & \multicolumn{1}{l}{Cost} & \multicolumn{1}{l}{Gap} & \multicolumn{1}{l|}{Time/s} & \multicolumn{1}{l}{Cost} & \multicolumn{1}{l}{Gap} & \multicolumn{1}{l}{Time/s} \\
    \midrule
    \midrule
    LS1   & 4.28  & 0.00\% & 331.45 & 4.86  & 0.00\% & 755.77 & 5.93  & 0.00\% & 2637.55 \\
    LS2   & 4.49  & 4.91\% & 362.28 & 5.16  & 6.17\% & 695.77 & 6.34  & 6.91\% & 2428.75 \\
    \midrule
    PN    & 16.25 & 279.67\% & 0.087 & 22.54 & 363.79\% & 0.117 & 29.68 & 400.51\% & 0.228 \\
    PN-dynamic & 6.47  & 51.17\% & \textbf{0.042} & 8.44  & 73.66\% & \textbf{0.076} & 11.17 & 88.36\% & 0.128 \\
    AM    & 6.11  & 42.76\% & 0.102 & 8.28  & 70.37\% & 0.162 & 10.52 & 77.40\% & 0.291 \\
    AM-dynamic & \textbf{5.19} & \textbf{21.26\%} & 0.071 & \textbf{6.01} & \textbf{23.66\%} & 0.082 & \textbf{7.62} & \textbf{28.50\%} & \textbf{0.106} \\
    \midrule
    AM-dynamic (LS) & \textbf{4.34} & \textbf{1.40\%} & \textbf{7.99} & \textbf{4.93} & \textbf{1.44\%} & \textbf{4.93} & \textbf{5.98} & \textbf{0.84\%} & \textbf{85.08} \\
    \bottomrule
    \bottomrule
    \end{tabular}%
  \label{tab:large}%
\end{table*}%

\begin{table*}[htbp]
  \centering
  \caption{Comparisons of the execution time when all the approaches reach a same cost.}
    \begin{tabular}{l|rr|rr|rr|rr|rr}
    \multicolumn{1}{c|}{\multirow{2}[1]{*}{Method}} & \multicolumn{2}{c|}{50-TSP} & \multicolumn{2}{c|}{100-TSP} & \multicolumn{2}{c|}{150-TSP} & \multicolumn{2}{c|}{200-TSP} & \multicolumn{2}{c}{300-TSP} \\
          & \multicolumn{1}{l}{Cost} & \multicolumn{1}{l|}{StopTime} & \multicolumn{1}{l}{Cost} & \multicolumn{1}{l|}{StopTime} & \multicolumn{1}{l}{Cost} & \multicolumn{1}{l|}{StopTime} & \multicolumn{1}{l}{Cost} & \multicolumn{1}{l|}{StopTime} & \multicolumn{1}{l}{Cost} & \multicolumn{1}{l}{StopTime} \\
    \midrule
    \midrule
    LS1   & 2.619 & 2.35  & 3.632 & 27.69 & 4.332 & 128.64 & 4.93  & 327.59 & 5.980  & 1295.65 \\
    LS2   & 2.642 & 11.14 & 3.742 & 71.43 & 4.476 & 290.66 & 5.086 & 559.48 & 6.266  & 2697.78 \\
    AM-dynamic (LS) & 2.641 & \textbf{0.78}  & 3.656 & \textbf{2.71}  & 4.337 & \textbf{8.07}  & 4.932 & \textbf{22.14} & 5.972 & \textbf{85.25} \\
    \bottomrule
    \bottomrule
    \end{tabular}%
  \label{tab:stop}%
\end{table*}%

\subsection{Validation of Generalization}
\label{general}
In this part, we analyze the generalization ability of the proposed approach to different types of CSP tasks. The models are trained on CSP instances with NC=7 (each city can cover its seven nearest neighbors), then are used to solve different types of CSP tasks. 

\begin{itemize}

\item We first generalize our model to CSP tasks with different NC. \tref{tab:nc} summarizes the results with NC=7, 11 and 15 respectively.

\item In the above CSP tasks, all cities cover the same number of nearby cities. In addition, we test the CSP task in which each city can cover different number of their nearby cities. To this aim, we generate CSP instances with random NCs ranging from 2 to 15 for each city. The original model trained on instances with NC=7 is used to test this task. \tref{tab:radius} presents the average results of 100 runs.

\item The same model is then used to solve another category of CSP: each city has a fixed coverage radius within that all other cities are covered. CSP tasks with a fixed coverage radius and variable coverage radius for each city are both tested. The coverage radius is set to a fixed value 0.2 for the first task, and the coverage radius is randomly chosen from a uniform distribution [0,0.25] for the second task. Results are reported in \tref{tab:radius}.

\end{itemize}

Results in \tref{tab:nc} demonstrate that the performance of our approach drops slightly when generalizing to CSP tasks with different NC values. The optimality gap exceeds 3\% and 2\% for test instances with NC=11 and NC=15 respectively. But our method is still able to produce near-optimal solutions with tiny optimality gap while requiring significantly less execution time compared to the traditional heuristic methods. It is noticed that a large optimality gap is found on CSP100 instances with NC=15. From the optimal cost we find it is similar with the results on CSP20 instances with NC=7, in which our model also performs the worst (6.94\%) when comparing to other test instances. It is therefore tentatively concluded that our proposed method performs worse when the length of the optimal tour is short. 

\begin{table*}[htbp]
  \centering
  \caption{Attention-dynamic method vs heuristic baselines on CSP tasks with different NCs. The average results of 100 runs are presented.}
    \begin{tabular}{c|l|rrr|rrr|rrr|rrr}
    \multirow{2}[3]{*}{NC} & \multicolumn{1}{c|}{\multirow{2}[3]{*}{Method}} & \multicolumn{3}{c|}{CSP100} & \multicolumn{3}{c|}{CSP150} & \multicolumn{3}{c|}{CSP200} & \multicolumn{3}{c}{CSP300} \\
\cmidrule{3-14}          &       & \multicolumn{1}{l}{Cost} & \multicolumn{1}{l}{Gap} & \multicolumn{1}{l|}{Time/s} & \multicolumn{1}{l}{Cost} & \multicolumn{1}{l}{Gap} & \multicolumn{1}{l|}{Time/s} & \multicolumn{1}{l}{Cost} & \multicolumn{1}{l}{Gap} & \multicolumn{1}{l|}{Time/s} & \multicolumn{1}{l}{Cost} & \multicolumn{1}{l}{Gap} & \multicolumn{1}{l}{Time/s} \\
    \midrule
    \midrule
    \multirow{3}[2]{*}{7} & LS1   & 3.58  & 0.00\% & 88.85 & 4.28  & 0.00\% & 331.45 & 4.86  & 0.00\% & 755.77 & 5.93  & 0.00\% & 2637.55 \\
          & LS2   & 3.69  & 3.07\% & 107.98 & 4.49  & 4.91\% & 362.28 & 5.16  & 6.17\% & 695.77 & 6.34  & 6.91\% & 2428.75 \\
          & AM-dynamic (LS) & 3.65  & 1.96\% & 2.66  & 4.34  & 1.40\% & 7.99  & 4.93  & 1.44\% & 4.93  & 5.98  & 0.84\% & 85.08 \\
    \midrule
    \midrule
    \multirow{3}[2]{*}{11} & LS1   & 2.94  & 0.00\% & 69.6  & 3.51  & 0.00\% & 230.38 & 4.01  & 0.00\% & 557.82 & 4.93  & 0.00\% & 1015.65 \\
          & LS2   & 3.03  & 3.06\% & 55.92 & 3.69  & 5.13\% & 144.77 & 4.22  & 5.24\% & 343.93 & 5.22  & 5.88\% & 810.56 \\
          & AM-dynamic (LS) & 3     & 2.04\% & 2.69  & 3.61  & 2.85\% & 5.92  & 4.15  & 3.49\% & 8.29  & 5.02  & 1.83\% & 28.45 \\
    \midrule
    \midrule
    \multicolumn{1}{c|}{\multirow{3}[2]{*}{15}} & LS1   & 2.58  & 0.00\% & 57.88 & 3.09  & 0.00\% & 170.93 & 3.53  & 0.00\% & 214.22 & 4.25  & 0.00\% & 1379.24 \\
          & LS2   & 2.64  & 2.33\% & 35.69 & 3.21  & 3.88\% & 96.37 & 3.65  & 3.40\% & 425.66 & 4.44  & 4.47\% & 720.74 \\
          & AM-dynamic (LS) & 3.02  & 17.05\% & 1.12  & 3.17  & 2.59\% & 4.22  & 3.6   & 1.98\% & 12.32 & 4.33  & 1.88\% & 35.26 \\
    \bottomrule
    \bottomrule
    \end{tabular}%
  \label{tab:nc}%
\end{table*}%

\begin{table*}[htbp]
  \centering
  \caption{Attention-dynamic method vs heuristic baselines on various CSP tasks. The average results of 100 runs are presented.}
    \begin{tabular}{c|l|rrr|rrr|rrr|rrr}
    \multirow{2}[3]{*}{CSP tasks} & \multicolumn{1}{c|}{\multirow{2}[3]{*}{Method}} & \multicolumn{3}{c|}{CSP50} & \multicolumn{3}{c|}{CSP100} & \multicolumn{3}{c|}{CSP150} & \multicolumn{3}{c}{CSP200} \\
\cmidrule{3-14}          &       & \multicolumn{1}{l}{Cost} & \multicolumn{1}{l}{Gap} & \multicolumn{1}{l|}{Time/s} & \multicolumn{1}{l}{Cost} & \multicolumn{1}{l}{Gap} & \multicolumn{1}{l|}{Time/s} & \multicolumn{1}{l}{Cost} & \multicolumn{1}{l}{Gap} & \multicolumn{1}{l|}{Time/s} & \multicolumn{1}{l}{Cost} & \multicolumn{1}{l}{Gap} & \multicolumn{1}{l}{Time/s} \\
    \midrule
    \midrule
    \multirow{3}[2]{*}{Varible NCs } & LS1   & 2.34  & 0.00\% & 5.62  & 2.96  & 0.00\% & 18.76 & 3.65  & 0.00\% & 93.25 & 4.14  & 0.00\% & 175.68 \\
          & LS2   & 2.36  & 0.85\% & 5.33  & 3.02  & 2.03\% & 22.53 & 3.77  & 3.29\% & 67.65 & 4.27  & 3.14\% & 128.65 \\
          & AM-dynamic (LS) & 2.54  & 8.55\% & 0.36  & 3.01  & 1.69\% & 1.56  & 3.7   & 1.37\% & 2.82  & 4.2   & 1.45\% & 6.52 \\
    \midrule
    \midrule
    \multirow{3}[2]{*}{Fixed radius } & LS1   & 2.94  & 0.00\% & 17.63 & 2.96  & 0.00\% & 49.1  & 2.93  & 0.00\% & 119.34 & 2.93  & 0.00\% & 237.54 \\
          & LS2   & 3.13  & 6.46\% & 21.4  & 3.04  & 2.70\% & 57.58 & 3.01  & 2.73\% & 87.78 & 2.96  & 1.02\% & 124.57 \\
          & AM-dynamic (LS) & 2.96  & 0.68\% & 1.52  & 2.99  & 1.01\% & 1.48  & 2.99  & 2.05\% & 2.42  & 2.97  & 1.37\% & 4.55 \\
    \midrule
    \midrule
    \multicolumn{1}{c|}{\multirow{3}[2]{*}{Varible radius }} & LS1   & 3.71  & 7.85\% & 17.84 & 3.41  & 0.59\% & 53.09 & 3.29  & 0.00\% & 81.28 & 3.17  & 0.63\% & 115.38 \\
          & LS2   & 3.76  & 9.30\% & 20.53 & 3.62  & 6.78\% & 47.24 & 3.56  & 8.21\% & 63.33 & 3.46  & 9.84\% & 85.21 \\
          & AM-dynamic (LS) & 3.44  & 0.00\% & 1.43  & 3.39  & 0.00\% & 9.56  & 3.31  & 0.61\% & 6.35  & 3.15  & 0.00\% & 3.45 \\
    \bottomrule
    \bottomrule
    \end{tabular}%
  \label{tab:radius}%
\end{table*}%

We can see from \tref{tab:radius} that our method can successfully generalize to various CSP tasks. 

\begin{figure}[htbp]
\centering
\subfloat[Attention-dynamic-LS (3.03)]{\includegraphics[width=1.7in]{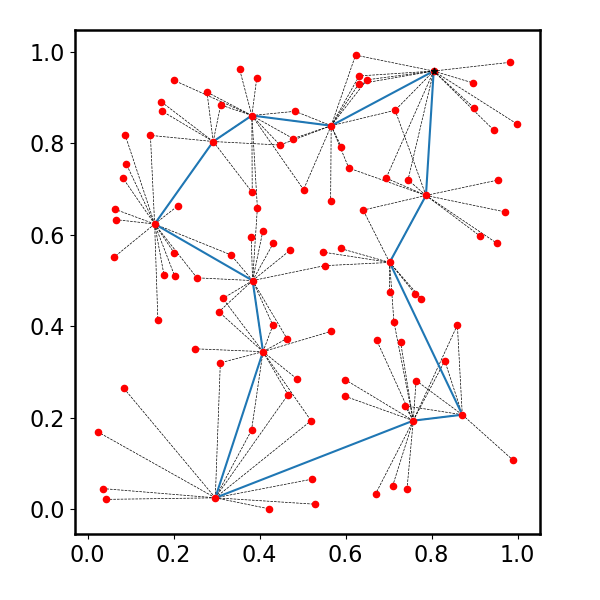}%
}
\hfil
\subfloat[LS1 (2.96)]{\includegraphics[width=1.7in]{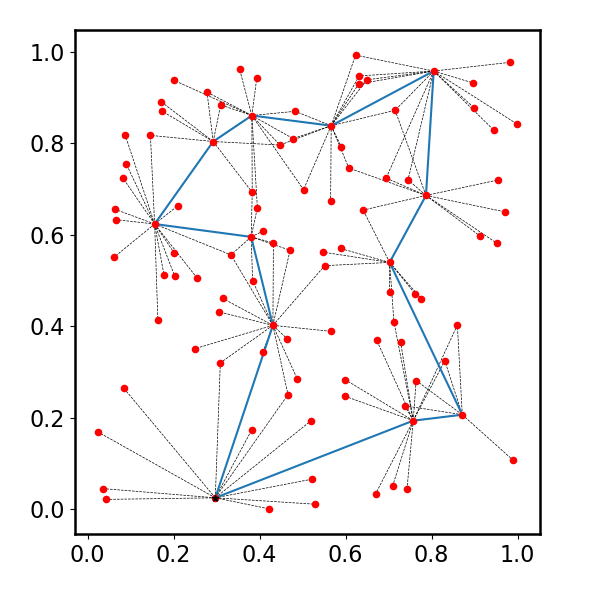}%
}
\caption{Solutions of a CSP100 instance where each city can cover a random number of cities. Tour lengths of the solutions are provided.}
\label{fig:nc}
\end{figure}

For CSP tasks in which different city can cover different number of cities, the model still shows a good performance. The optimality gap of our method can be always within 2\% for all CSP instances except CSP50. The worse performance of our approach on CSP50 is consistent with the above experiments. \fref{fig:nc} visualizes the solutions obtained by our method and LS1 on a CSP100 instance with randomly generated NC values. There are only two different cities between the solutions of the two methods, leading to a slight worse performance of our method than LS1. 

\begin{figure}[htbp]
\centering
\subfloat[Attention-dynamic-LS (3.28)]{\includegraphics[width=1.7in]{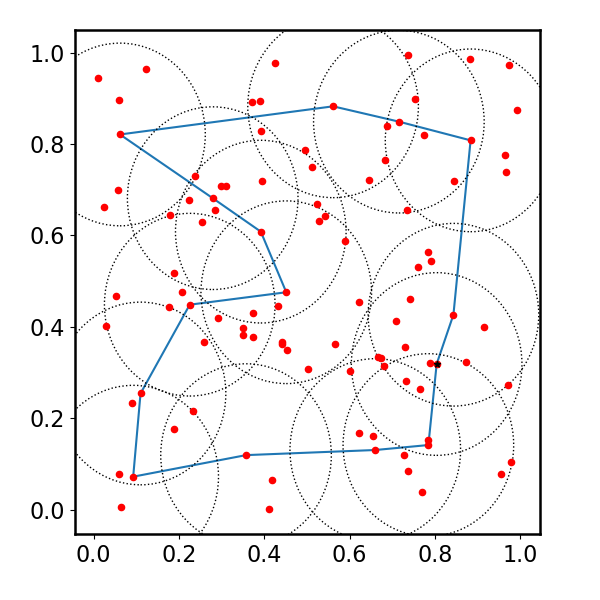}%
}
\hfil
\subfloat[LS1 (3.16)]{\includegraphics[width=1.7in]{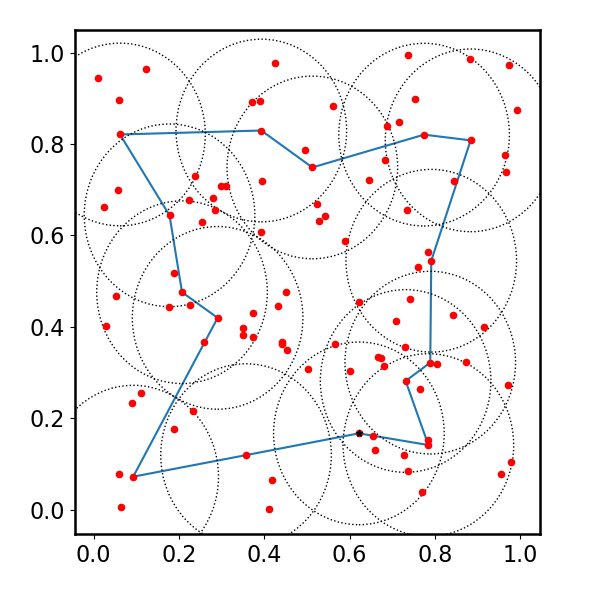}%
}
\caption{Solutions of a CSP100 instance with a fixed cover radius for each city. Tour lengths of the solutions are provided.}
\label{fig:radius}
\end{figure}

\begin{figure}[htbp]
\centering
\subfloat[Attention-dynamic-LS (3.26)]{\includegraphics[width=1.7in]{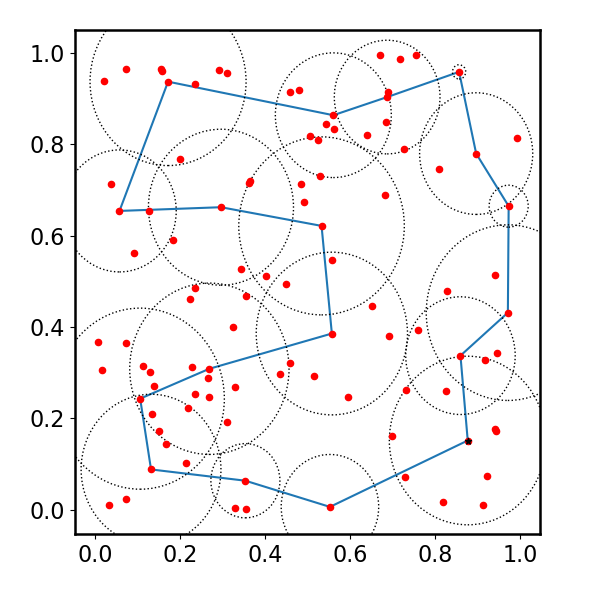}%
}
\hfil
\subfloat[LS1 (3.41)]{\includegraphics[width=1.7in]{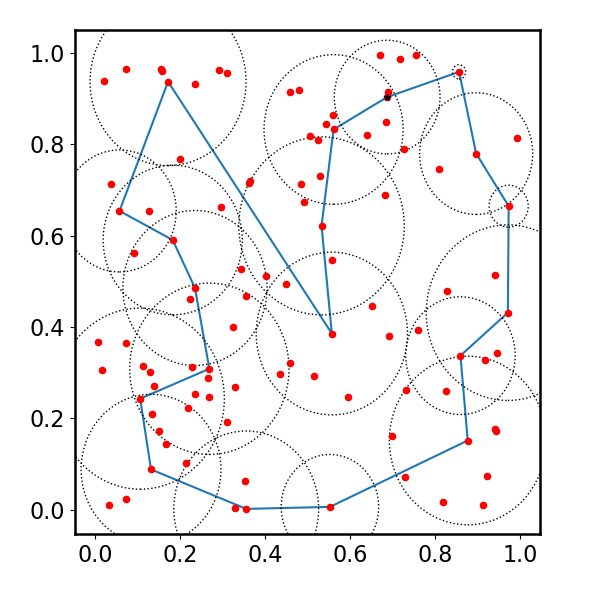}%
}
\caption{Solutions of a CSP100 instance where each city has a random covering radius. Tour lengths of the solutions are provided.}
\label{fig:radius2}
\end{figure}

Notably, our method outperforms LS1 on CSP tasks with variable cover radius. Our method also achieves a closer optimality to LS1 on CSP instances with fixed cover radius. Although the model is not trained on these type of CSP tasks, it can still tackle them successfully. \fref{fig:radius} and \fref{fig:radius2} visualize the solutions on CSP instances with fixed and variable cover radius respectively. It can be seen that the obtained solutions of the two compared methods are similar.

It is obvious that our approach can always produce a comparable solution while requiring significantly less execution time. The model only trained on CSP instances of NC=7 can be used to tackle various types of CSP tasks. This generalization is mainly achieved by the dynamic embeddings that we design. Once a city is covered, its probability of being selected is dynamically adjusted by the learned model parameters. 

\section{Conclusion}
Covering salesman problems arise in numerous application domains such as the disaster planning. However, with the problem dimension increasing in real-world applications, traditional approaches suffer from the limitation of forbidding execution time. Moreover, considerable development efforts are required to design the heuristics. In this context, we introduce a new deep learning approach for the CSP in this study. Different from the  traditional solvers, in this approach, a deep neural network is used to directly output a permutation of the input. The model is trained using the REINFORCE algorithm with greedy rollout in an unsupervised manner. As an end-to-end approach, this method shows desirable properties of fast solving speed and the ability of generalizing to unseen instances. There is still an optimality gap of our method with respect to traditional solvers. However, in some occasions where the problem is large-scale and requires quick decision, the proposed approach is highly desirable as it requires significantly less execution time. Moreover, esoteric domain knowledge and trial-and-error process is needed to design traditional heuristic methods. Years of effort has been devoted to engineer the heuristic rules for CSPs, however, no actual breakthrough is accomplished. Most of the designed heuristics are problem-specific and need to be revised once the problem setting changes. Thus the deep learning approach proposed in this study can be more favorable as it can learn the heuristic from the data by itself and automate the process of decision-making. 

The proposed method provides a desirable trade-off of the solution quality and the execution time. It also significantly outperforms the current state-of-the-art deep learning approaches for combinatorial optimization. However, the better solution quality is still the significant advantage of traditional solvers. Improving the solution quality of the deep learning approach is an urgent research perspective in the future. The main challenge is how to handle the dynamic feature of the problem more effectively. Better mechanism should be developed to enable the agent to understand the change of the context when a city is visited and others are covered, and how this dynamic change affects the policy.


%



\ifCLASSOPTIONcaptionsoff
  \newpage
\fi



\bibliographystyle{IEEEtran}
\bibliography{IEEEfull}




\end{document}